\documentclass{article}

\usepackage{microtype}
\usepackage{graphicx}
\usepackage{subfigure}
\usepackage{booktabs} 

\usepackage{hyperref}

\usepackage[accepted]{icml2025}

\usepackage{amsmath}
\usepackage{amssymb}
\usepackage{mathtools}
\usepackage{amsthm}

\usepackage[capitalize,noabbrev]{cleveref}

\theoremstyle{plain}

\theoremstyle{definition}

\theoremstyle{remark}

\usepackage{multicol,multirow}
\usepackage{makecell}
\usepackage{adjustbox}
\usepackage{hyphenat}
\usepackage{tikz}
\usepackage{wrapfig}
\usepackage{enumitem}
\usepackage{array}
\usepackage{arydshln}

\usepackage[textsize=tiny]{todonotes}

\icmltitlerunning{Generalising Battery Control in Net-Zero Buildings via Personalised Federated RL}

\newcommand{\nl}{\\[2pt]}

\begin{document}

\twocolumn[
\icmltitle{Generalising Battery Control in Net-Zero Buildings via Personalised Federated RL}



\icmlsetsymbol{equal}{*}

\begin{icmlauthorlist}
    \icmlauthor{Nicolas Cuadrado}{equal,yyy}
    \icmlauthor{Samuel Horváth}{yyy}
    \icmlauthor{Martin Takáč}{yyy}
\end{icmlauthorlist}

\icmlaffiliation{yyy}{Department of Machine Learning, Mohamed Bin Zayed University of Artificial Intelligence, Abu Dhabi, UAE}

\icmlcorrespondingauthor{Nicolas Cuadrado}{nicolas.avila@mbzuai.ac.ae}

\icmlkeywords{Net-zero, Federated Learning, Personalization, Reinforcement Learning}

\vskip 0.3in
]



\printAffiliationsAndNotice{\icmlEqualContribution} 

\begin{abstract}
This work studies the challenge of optimal energy management in building-based microgrids through a collaborative and privacy-preserving framework. We evaluated two common RL algorithms (PPO and TRPO) in different collaborative setups to manage distributed energy resources (DERs) efficiently. Using a customized version of the CityLearn environment and synthetically generated data, we simulate and design net-zero energy scenarios for microgrids composed of multiple buildings. Our approach emphasizes reducing energy costs and carbon emissions while ensuring privacy. Experimental results demonstrate that Federated TRPO is comparable with state-of-the-art federated RL methodologies without hyperparameter tuning. The proposed framework highlights the feasibility of collaborative learning for achieving optimal control policies in energy systems, advancing the goals of sustainable and efficient smart grids. Our code is accessible in \href{https://github.com/Optimization-and-Machine-Learning-Lab/energy_fed_trpo.git}{\textit{this repo}}.

\end{abstract}

\section{Introduction}
\label{introduction}

Global reviews of progress under the SDGs and the 2015 Paris Agreement still show the world drifting beyond the 1.5–2 °C ceiling, as confirmed by the latest IPCC assessment \cite{IPCCAR6S8:online}. Because building operations consume about 30\% of global final energy and produce roughly 26\% of energy-related CO$_2$ emissions \cite{IEA_Buildings}, cutting their footprint is pivotal. A recent analysis estimates that AI-enabled controls alone could trim building energy demand and emissions by 8–19\% by 2050 \cite{AI_Building_nature}. Machine learning (ML) is a technology that, when used correctly, can accelerate the actions needed to fight global warming. ML can support the transition from fossil fuels, accelerate renewable energy adoption, improve energy efficiency, and promote sustainable transportation, offering powerful tools to meet some of the most pressing climate goals \cite{DBLP:journals/corr/abs-1906-05433}.
\nl
Rooftop solar and other on-site renewables introduce fast power swings that can create acute shortage events at the building level \cite{renewable_shortage_nature2024,climate_change_grid, climate_impact_renewables}. Modern Building Energy Management Systems (BEMS) mitigate these fluctuations by pairing local storage with predictive control and have been shown to cut energy use and costs across diverse facilities \cite{BEMS_review_2024}. Model-free reinforcement learning (RL) can learn such control policies directly from data, capturing the non-linear dynamics of distributed resources without a detailed physics model \cite{rl_microgrid_survey}. Yet standard RL is sample-hungry and generalises poorly to unseen climates or occupancy patterns \cite{rl_generalisation}. Privacy-preserving federated approaches overcome both issues by exchanging model updates rather than raw traces, protecting occupant data while pooling experience across sites \cite{fl_privacy}. Recent demonstrations report cost and CO$_2$ reductions when scaling federated RL to dozens of heterogeneous buildings in multiple climates \cite{frl_netzero}. Privacy guarantees remove regulatory and organisational barriers to pooling knowledge across many buildings, making collaborative learning a practical path to scalable net-zero control \cite{frl_scaling}.
\nl
This study explores building-scale net-zero microgrids (``a group of interconnected loads and distributed energy resources within clearly defined electrical boundaries that acts as a single controllable entity'' \cite{us_doe_microgrids}) as test-beds for scalable, privacy-preserving control. We generated scenarios where buildings are self-sustainable by design as long as the electrical storage is managed using the optimal policy. Using a custom microgrid scenario based on CityLearn \cite{DBLP:conf/sensys/Vazquez-Canteli19}, we explored the generalization capabilities of RL agents and the challenges they faced even in simplified scenarios showing that naive policies are difficult to learn in a collaborative way. We improved the learning process using Federated Learning (FL), in this setup each building trains a local RL agent but only shares parameter updates, a practice that retains data privacy while exploiting the scalability and generalisation benefits identified in recent federated-learning surveys \cite{Kairouz2021Advances,fl_privacy}. Additionally, we included personalisation through split-learning \cite{DBLP:journals/corr/abs-2212-08343}, further shrinking the generalisation gap that often appears when controllers are deployed in heterogeneous conditions or occupancy patterns \cite{rl_generalisation}. We present a summary of our contributions next:
\begin{itemize}[noitemsep,topsep=0pt]

    \item We introduce a custom environment based on CityLearn, which allows a better understanding of the learned policies and the impact of the reward function in the learning process. We also modified some characteristics that better represent the real-life dynamics.

    \item We provide experimental assessment of traditional RL algorithms in our environment, testing generalization and optimality of the learned policies.
    
    \item Our results demonstrate the generalization capabilities of FL and how reasonable considerations like personalization and split-learning \cite{DBLP:journals/corr/abs-2212-08343} can reduce the optimality gap in the learned policies.
    
    \item We provide a wrapper for the CityLearn environment that makes it compatible with TorchRL \cite{bou2023torchrldatadrivendecisionmakinglibrary}, improving the modularity and portability of our code. We also provided a version of the TRPO algorithm using TorchRL.
    
\end{itemize}
\section{Related Work}

{\bf Federated Learning.} FL trains a common model by letting each client update locally and then merging weights. The seminal FedAvg algorithm performs this aggregation on a central server \cite{DBLP:conf/aistats/McMahanMRHA17}, whereas Gossip protocols average peer-to-peer without a server \cite{DBLP:conf/icml/KoloskovaSJ19}. Subsequent work addresses FedAvg’s limits under non-IID data and privacy leakage: split learning keeps part of each network private to enable personalisation \cite{DBLP:journals/corr/abs-1812-00564}, and differential privacy injects calibrated noise into updates \cite{DBLP:journals/corr/abs-1712-07557}. Studies on heterogeneous clients \cite{DBLP:conf/iclr/Diao0T21} and generalisation theory \cite{DBLP:conf/iclr/0002MNS22} confirm that, with these extensions, FL can learn robust, privacy-aware controllers—making it well suited to smart-grid scenarios with diverse buildings.
\nl
{\bf RL-based and Federated RL-based microgrids.} Recent studies model microgrids as groups of houses with distinct loads, rooftop renewables, and storage devices, typically using the open-source CityLearn platform \cite{DBLP:conf/sensys/Vazquez-Canteli19,Nweye_2024} built on Gymnasium \cite{Towers_Gymnasium_A_Standard} and real demand data \cite{oedi_4520}. Within CityLearn, \cite{nweye2023merlin} trained multi-agent Soft Actor-Critic controllers for battery systems to cut cost, carbon, and peak demand, while \cite{nweye2022real} showed that warm-starting RL with rule-based policies speeds convergence. Beyond public benchmarks, a model-based RL approach with constrained optimisation targets multi-energy hubs \cite{DBLP:journals/corr/abs-2104-09785}; a federated RL scheme addresses free-riders and non-IID data \cite{DBLP:journals/tii/SuWLZLCC22}; and hierarchical FRL embeds FedAvg in layered control architectures \cite{DBLP:conf/iclr/CuadradoGT23,DBLP:journals/corr/abs-2303-08447}. Collectively, these works highlight both the promise and the outstanding challenges of RL and FRL for collaborative, net-zero microgrids.
\nl
Despite steady progress, two gaps remain. First, many studies treat a monotonic rise in cumulative reward as proof of optimality, yet trivial local minima, such as always charging or doing nothing, can produce similar curves. Second, CityLearn challenge winners (2021–2023) optimised the highest-weighted terms and largely ignored CO$_2$ impact \cite{nweye2022citylearn,10.1145/3486611.3492226}, even though emission reduction should be central in climate-aligned control.
\section{Method} \label{sec:method}

\subsection{Environment} \label{subsec:env}

We used our customized version of the CityLearn environment. We decided to generate synthetic data that allows us to understand clearly the learned policies through simple cases, focusing on representing homes that can be self-sustainable, in other words, each home can provide enough energy to serve its demand, so the optimal solution must have zero cost and emissions. Our data generation process takes as a reference the \textit{citylearn\_challenge\_2022\_phase\_all} scenario. It modifies the time series for \textit{non-shiftable load} to ensure solar energy can cover the demand profile completely. We included other modifications to the environment:
\begin{itemize}[noitemsep,topsep=0pt]
    
    \item New logic for the selling energy rate. We found it unrealistic that in the original scenario, the users receive compensation equal to the Time-of-Use (ToU) tariff. This setup discourages houses from being self-sustainable.

    \item We simplified the storage models so they don't suffer degradation or decay in their performance, helping to ease the understanding of the policies we are learning, which is one of the main interests of this research.
    
    \item We introduced additional observations representing load predictions for some steps ahead.
    
\end{itemize}
We kept the grid price pattern in the original CityLearn environment, which fluctuates throughout the day, peaking during high-demand hours, while the solar panels contribute energy during daylight hours. Allowing the agent to stockpile energy during off-peak times and, when solar generation exceeds immediate needs, later release stored energy during high-price periods. Emissions follow the actual emission rate from the grid mix (kgCO2e/kWh) following the original dataset used in CityLearn. We generate convex combinations of the training buildings' time series for the validation data. Each episode spans twenty-four hours, and $t$ is the index for the time step. For a building $i$, an agent considers the accessible observations to decide the charge or discharge of the battery in a time step ($E^{\text{batt}}_{t,i}$), aiming to meet the building's load requirements. Its primary objective is minimizing its house's overall energy cost and emissions and, as an extension, for the whole microgrid. The table \ref{tab:rl_case} in the Appendix presents an overview of the RL case. Our design of the reward function is a weighted sum of the resulting cost and emissions for the complete episode (a full day), and we established the values as $W_C = 0.4$ and $W_G = 0.6$ to enforce the prioritization of emission reduction. We also added a penalization to the reward function to guide the agent to avoid unfeasible actions. 
\nl
\begin{figure}[ht]
    \centering
    \includegraphics[width=\linewidth]{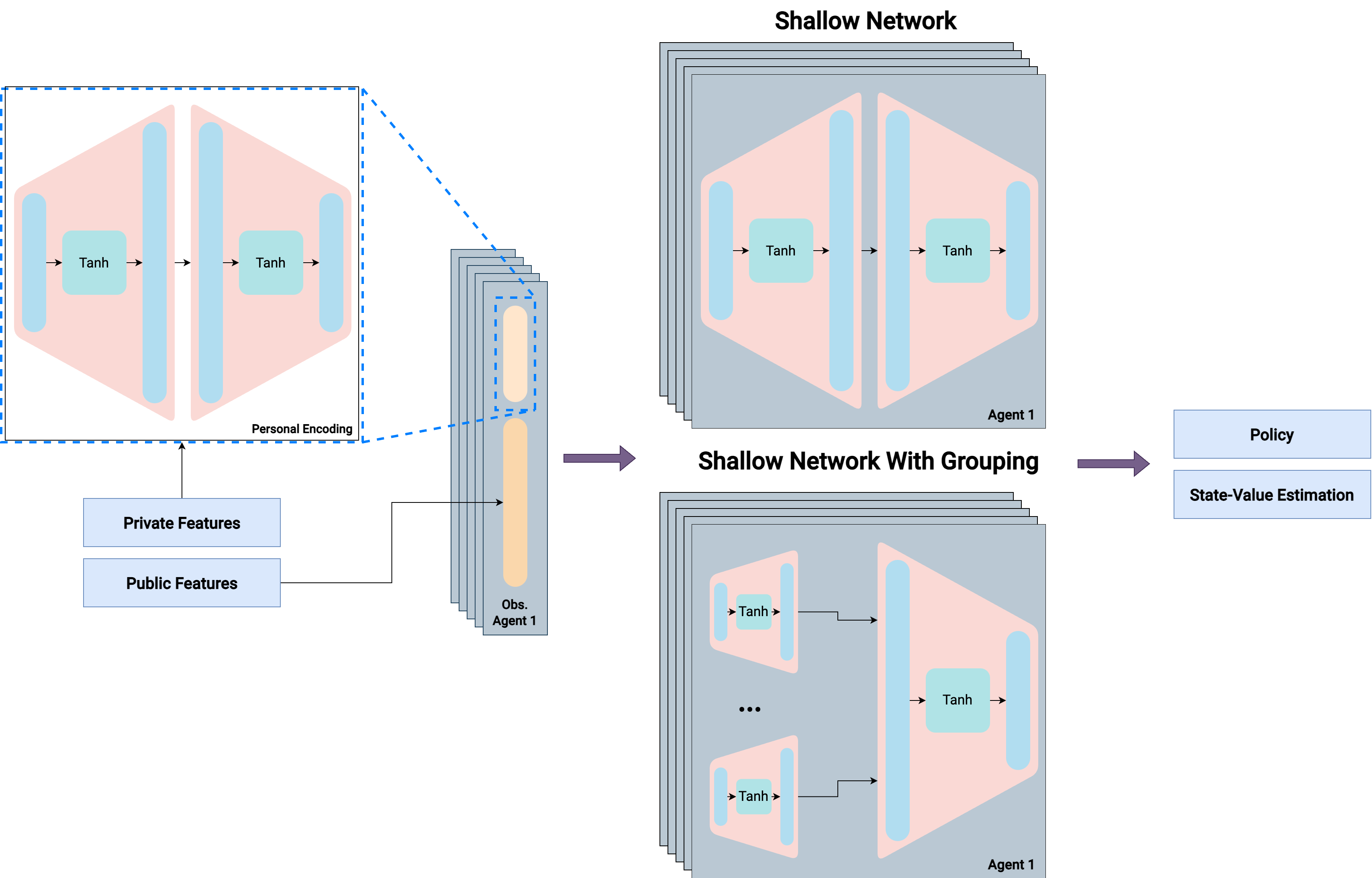}
    \caption{Overview of our models: We use the same type of network for Policy and State-Value networks, changing only the output layer. The \textit{Personal Encoding} block is optional.}
    \label{fig:model}
    \vspace{-10mm}
\end{figure}

\subsection{Model}

We explore the advantages of a collaborative learning approach for the case of microgrids, given their distributed nature: each household has an agent that explores and learns from different sections of the state space, given the individual attributes of the household (generation, load profile, battery capacity). We have two types of models for both Policy and Critic networks: a shallow neural network and a custom model that adds a grouping feature logic as the first part of the feature extraction; we refer to it as \textit{Personal Encoding} in Figure \ref{fig:model}. This block represents a part of the network that remains local; the FedAvg process does not access it. It encodes features that should remain private (e.g., number of people living in the household, ToU strategy, appliances, area of the house, etc.) and learns a latent representation, used as an input to the public model, stacked with the public features, which aligns with the concept of split-learning \cite{DBLP:journals/corr/abs-1812-00564}. We can assume it as a personalization strategy.
\section{Experiments}

\begin{table*}[ht]
    \centering
    \caption{\textbf{Experimental Results for 2-Building Environment:} Values in the table represent the average for $5$ seeds. Values in parentheses represent the difference concerning not having battery management.}\label{tab:exp_results_2b}
    \resizebox{\textwidth}{!}{%
        \begin{tabular}{llcccccc}
        \hline
        \multirow{2}{*}{\textbf{}} & \textbf{Experiment} & 
        \makecell{\textbf{Train} \\ \textbf{Cost}} & 
        \makecell{\textbf{Eval} \\ \textbf{Cost}} & 
        \makecell{\textbf{Train} \\ \textbf{Emissions}} & 
        \makecell{\textbf{Eval} \\ \textbf{Emissions}} & 
        \makecell{\textbf{Train} \\ \textbf{Reward}} & 
        \makecell{\textbf{Eval} \\ \textbf{Reward}} \\
        \hline
        \multirow{8}{*}{\textbf{PPO}}
        & base & 0.04227	(-0.08977) & 0.04099 (-0.09148) & 0.06380 (-0.07487) & 0.06260 (-0.07652) & -0.19331 & -0.18754 \\
        & \textbf{gf} & 0.03054 (-0.10150) & \textbf{0.03049 (-0.10199)} & 0.04341 (-0.09526) & \textbf{0.04308 (-0.09604)} & -0.13227 & \textbf{-0.13188} \\
        & pe & 0.03885 (-0.09319) & 0.04630 (-0.08618) & 0.05841 (-0.08026) & 0.06384 (-0.07527) & -0.17686 & -0.19956 \\
        & \textbf{pe gf} & \textbf{0.03051	(-0.10153)} & 0.03227 (-0.10021) & \textbf{0.04306 (-0.09561)} & 0.04637 (-0.09274) & \textbf{-0.13065} & -0.14021 \\
        \cdashline{2-8}[.4pt/1pt]
        & \textbf{base shifted} & 0.04923	(-0.09717) & \textbf{0.07062 (-0.07229)} & 0.08726 (-0.10790) & \textbf{0.11533 (-0.08125)} & -0.22554 & \textbf{-0.34312} \\
        & gf shifted & 0.24691 (+0.10051) & 0.25710 (+0.11419) & 0.28074 (+0.08558) & 0.30601 (+0.10943) & -1.46458 & -1.51916 \\
        & \textbf{pe shifted} & \textbf{0.04541 (-0.10099)} & 0.10654 (-0.03637) & \textbf{0.08211 (-0.11305)} & 0.13600 (-0.06058) & \textbf{-0.21291} & -0.56921 \\
        & pe gf shifted & 0.10324	(-0.04316) & 0.15339 (+0.01049) & 0.10922 (-0.08594) & 0.17679 (-0.01979) & -1.10813 & -1.34919 \\
        \hline
        \multirow{8}{*}{\textbf{TRPO}}
        & base & 0.31658	(+0.18454) & 0.32391 (-0.19143) & 0.26502 (+0.12635) & 0.26908 (+0.12997) & -1.12509 & -1.12524 \\
        & gf & 0.31882 (+0.18678) & 0.32288 (+0.19040) & 0.27001 (+0.13134) & 0.27020 (+0.13108) & -1.02230 & -1.01532 \\
        & \textbf{pe} & 0.10567 (-0.02637) & 0.10402 (-0.02846) & \textbf{0.11511 (-0.02357)} & \textbf{0.11271 (-0.02641)} & \textbf{-0.47736} & \textbf{-0.47064} \\
        & \textbf{pe gf} & \textbf{0.10519 (-0.02685)} & \textbf{0.10355 (-0.02893)} & 0.11519 (-0.02348) & 0.11302 (-0.02609) & -0.47877 & -0.47164 \\
        \cdashline{2-8}[.4pt/1pt]
        & base shifted & 0.15482 (+0.00842) & 0.14608 (+0.00317) & 0.18679 (-0.00837) & 0.17856 (-0.01802) & -0.61017 & -0.58478 \\
        & gf shifted & 0.17848 ( 0.03208) & 0.16951 (+0.02660) & 0.19592 (+0.00076) & 0.18621 (-0.01037) & -0.77824 & -0.74692 \\
        & pe shifted & 0.14833 (+0.00193) & 0.14171 (-0.00120) & 0.17103 (-0.02413) & 0.16301 (-0.03357) & -0.64225 & -0.63830 \\
        & \textbf{pe gf shifted} & \textbf{0.12728 (-0.01912)} & \textbf{0.12032 (-0.02259)} & \textbf{0.15287 (-0.04229)} & \textbf{0.14513 (-0.05145)} & \textbf{-0.57360} & \textbf{-0.56441} \\
        \hline
        \end{tabular}%
    }
    \caption{\textbf{Experimental Results for 5-Building Environment:} Values in the table represent the average for $5$. Values in parentheses represent the difference concerning not having battery management.}\label{tab:exp_results_5b}
    \resizebox{\textwidth}{!}{%
        \begin{tabular}{llcccccc}
        \hline
        \multirow{2}{*}{\textbf{}} & \textbf{Experiment} & 
        \makecell{\textbf{Train} \\ \textbf{Cost}} & 
        \makecell{\textbf{Eval} \\ \textbf{Cost}} & 
        \makecell{\textbf{Train} \\ \textbf{Emissions}} & 
        \makecell{\textbf{Eval} \\ \textbf{Emissions}} & 
        \makecell{\textbf{Train} \\ \textbf{Reward}} & 
        \makecell{\textbf{Eval} \\ \textbf{Reward}} \\
        \hline
        \multirow{8}{*}{\textbf{PPO}}
        & base & 0.04001	(-0.09140) & 0.03856 (-0.09297) & 0.06094 (-0.07710) & 0.05949 (-0.07866) & -0.18392 & -0.17930 \\
        & gf & 0.24239 (+0.11098) & 0.25020 (+0.11867) & 0.23766 (+0.09962) & 0.24276 (+0.10461) & -1.55780 & -1.55296 \\
        & pe & 0.04089 (-0.09052) & 0.04505 (-0.08648) & 0.06159 (-0.07645) & 0.06488 (-0.07328) & -0.18951 & -0.20773 \\
        & \textbf{pe gf} & \textbf{0.02889	(-0.10253)} & \textbf{0.03233 (-0.09920)} & \textbf{0.04235 (-0.09569)} & \textbf{0.04711 (-0.09104)} & \textbf{-0.13100} & \textbf{-0.14662} \\
        \cdashline{2-8}[.4pt/1pt]
        & base shifted & 0.05985	(-0.09039) & 0.06716 (-0.08331) & 0.08469 (-0.07876) & 0.09063 (-0.07437) & -0.26297 & -0.28186 \\
        & \textbf{gf shifted} & \textbf{0.03725 (-0.11299)} & \textbf{0.05633 (-0.09414)} & \textbf{0.05421 (-0.10924)} & \textbf{0.07025 (-0.09474)} & \textbf{-0.16773} & \textbf{-0.22911} \\
        & pe shifted & 0.06305 (-0.08719) & 0.08847 (-0.06200) & 0.09080 (-0.07265) & 0.10826 (-0.05674) & -0.28214 & -0.39795 \\
        & pe gf shifted & 0.10324	(-0.04316) & 0.15339 (+0.01049) & 0.10922 (-0.08594) & 0.17679 (-0.01979) & -1.10813 & -1.34919 \\
        \hline
        \multirow{4}{*}{\textbf{TRPO}}
        & \textbf{base} & \textbf{0.09811 (-0.03330)} & \textbf{0.09737 (-0.03416)} & 0.11173 (-0.02630) & 0.11086 (-0.02730) & \textbf{-0.46143} & -0.45768 \\
        & gf & 0.10437 (-0.02704) & 0.10312 (-0.02841) & 0.11455 (-0.02349) & 0.11327 (-0.02488) & -0.47631 & -0.47153 \\
        & \textbf{pe} & 0.09843 (-0.03298) & 0.09750 (-0.03403) & \textbf{0.11019 (-0.02785)} & \textbf{0.10906 (-0.02909)} & -0.46143 & \textbf{-0.45614} \\
        & pe gf & 0.10177	(-0.02964) & 0.10044 (-0.03109) & 0.11274 (-0.02529) & 0.11140 (-0.02676) & -0.46460 & -0.45958 \\
        \cdashline{2-8}[.4pt/1pt]
        & \textbf{base shifted} & 0.13025	(-0.01999) & \textbf{0.11929 (-0.03118)} & \textbf{0.12936 (-0.03409)} & \textbf{0.12229 (-0.04270)} & -0.57766 & \textbf{-0.53850} \\
        & gf shifted & 0.17795 (+0.02771) & 0.16457 (+0.01410) & 0.16254 (-0.00091) & 0.15295 (-0.01204) & -0.77981 & -0.74353 \\
        & \textbf{pe shifted} & \textbf{0.12846 (-0.02178)} & 0.11979 (-0.03068) & 0.13076 (-0.03269) & 0.12391 (-0.04109) & \textbf{-0.56240} & -0.55708 \\
        & pe gf shifted & 0.14833	(+0.00193) & 0.14171 (-0.00120) & 0.17103 (-0.02413) & 0.16301 (-0.03357) & -0.64225 & -0.63830 \\
        \hline
        \end{tabular}%
    }
\end{table*}

\subsection{Experiment Setup}

We generated two environments for our experiments using the simplified version of the data \ref{fig:simple_solar_load}: a microgrid with two houses and one with five houses. Each one also has a shifted version that creates day gaps between the house's time series, simulating that each house is subject to different environmental conditions and solar irradiance patterns. Regarding the model configurations, as described \ref{fig:model}, we run experiments in four setups: shallow network, shallow network with personal encoding, shallow network with grouping, and shallow network with personal encoding and grouping. We defined four group features: load-related, solar-related, pricing-related, and the rest. We fixed the communication rounds for both algorithms we tried, Proximal Policy Optimisation(PPO) and Trust Region Policy Optimisation (TRPO), to evaluate how close they can come to the optimal solution within the same number of iterations. Another essential detail is setting the discount factor $\gamma = 1$, meaning that our agents don't consider a discounted reward, given that we are running episodes of just one day (24 time steps).

\subsection{Quantitative Results}

The tables \ref{tab:exp_results_2b} and \ref{tab:exp_results_5b} show the experiments we mentioned in the previous section: \textit{base} means we used the shallow network, \textit{pe} that we included personal encoding, and \textit{gf} that we grouped features. We ran each combination for five different seeds. We averaged the results, making $160$ experiments with an average run time of $70m$ for the PPO runs, and $110m$ for the TRPO runs using an NVIDIA RTX 4000 Ada Generation GPU.
\nl
Starting with the environment containing just two buildings, we found that training with the shifted variant is more challenging than the normal one, which makes sense considering we provide more diverse data without increasing the minibatch size. We cannot conclude from the results that using personal encoding or grouped features made a difference in the case of PPO, but for TRPO, it seems that they could be beneficial, as was evident in the shifted dataset. In general, as expected, the performance of the tuned PPO was superior to that of the TRPO, and the reward function value demonstrated it.
\nl
In the five-building environment, grouping features seems to play a more critical role than in the case of just having two buildings. We interpret this as the effect of having more agents, which means more sampling and more diverse data, which can help the networks unveil patterns in the data that are not evident by simple inspection. However, this does not seem to be the case for TRPO, in which the simple model yields the best performance, although it is essential to note that the performance of TRPO is poor based on the reward function. PPO overperforms TRPO significantly, but we highlight that we didn't spend time fine-tuning it. Still, it managed to identify some patterns in the data, as we will show in the qualitative results in the Appendix.

\section{Conclusion}

We examined battery control in net-zero microgrids. Using a stripped-down CityLearn scenario with a known optimal policy, we studied RL combined with FedAvg and split-learning personalisation to balance privacy, heterogeneity, and policy quality. Federated TRPO performed on par with a tuned federated PPO: PPO was sometimes superior but stalled near the optimum, whereas TRPO converged reliably without hyperparameter tuning. A pragmatic remedy is to warm-start with a few PPO iterations and then switch to TRPO, leveraging TRPO’s adaptive step size in the final optimisation phase. The study shows that even ``simple'' optimal policies are surprisingly hard to learn, supporting curriculum designs that start with well-controlled cases before moving to full real-world complexity.
\nl
\textbf{Future Work.} Next steps include benchmarking against model-predictive control (MPC) and additional RL baselines to gauge relative sample-efficiency and robustness. We will design new ``toy'' microgrid cases—varying storage sizes, demand shapes, and renewable mixes—to probe policy behaviour and measure the optimality gap more precisely. Finally, we plan to scale experiments from five to dozens of buildings, testing how privacy, communication cost, and convergence behave in larger federations.


\newpage
\bibliography{ref}
\bibliographystyle{icml2025}

\newpage
\appendix
\onecolumn

\section{Definition of RL Scenario}

\subsection{Reinforcement Learning}

The RL component of our methods happens at the local training. We compared TRPO against a state-of-the-art algorithm called Proximal Policy Optimization (PPO) \cite{schulman2017proximalpolicyoptimizationalgorithms}, which emerged as an alternative to TRPO that avoids the use of second-order information. PPO relies on the same theoretical base as TRPO but redefines the policy optimization objective as an unconstrained expression. The scenario we study in this work, and the case of microgrids' agents, does not require the use of very deep neural networks, and that is a reason why we believe that second-order methods are feasible in these cases, especially for RL methods, which tend to be very sensitive to the learning rate. TRPO explores the vast space by identifying suitable step sizes within a confined region, ensuring stable policy, monotonic improvement, and more effective exploration than other policy gradient methods.

Another essential aspect of opting for an RL approach is defining a proper reward function, especially considering that we defined a continuous action space, which means the number of possible actions is unbounded. The typical approach for this case is to use a policy network that outputs the parameters of a Gaussian distribution (reparameterization trick) to sample a real value that defines the action from that distribution. Throughout the training process, we don't have guarantees that the policy network will explore feasible actions (within the range specified in \ref{tab:rl_case}), especially in the first iterations. That fact motivated the addition of a penalization $P_{i,t}$ that impacts the return of a trajectory (the accumulated sum or rewards.) An essential consideration we made when designing our reward function is that we are not guiding it to follow a specific policy we know beforehand; we are just guiding the exploration of the agent by avoiding ineffective actions.

\subsection{Federated RL}

An actual FL setup implies having clients interact through a communication network channel throughout the training rounds. In the context of our study case, we can imagine it as the agents operating in an embedded device that controls the batteries of each household and connects to a parameter server through a dedicated network connection. However, this case would introduce some of the challenges of FL: limited bandwidth, faulty nodes or updates, Byzantine clients, etc. The main interest of this work is not to alleviate those challenges but to explore the benefits of collaborative learning. Thus, we are not required to implement the exact flow of FL to perform our experiments.

As Figure \ref{fig:model} shows, we stack the features (observations and personal encoding) for all the agents and pass them to their corresponding networks. Passing the stacked observations to a single network has the same effect as doing a FedAvg: each agent samples their trajectories, and after a forward pass, the backward pass will propagate the average gradient of all of them, which is, in practice, the same behavior we expect from the FL methodology. We use a simple one-hot encoder for each home as the personal encoding. Although we are not in the process of training a private part of the policy network, we can explore the exact effects of personalization by having a vector representation that identifies each home. Using FL is to discern the interconnected relationships in the observations across buildings collectively. Relying solely on data from a single building risks overfitting, but leveraging diverse data distributions from multiple buildings helps uncover latent connections. Moreover, the model adeptly extrapolates unseen data instances by assimilating insights from disparate building data distributions, handling anomalies emerging from using usage patterns.

\begin{figure}[htb]
    \centering
    \includegraphics[width=0.8\linewidth]{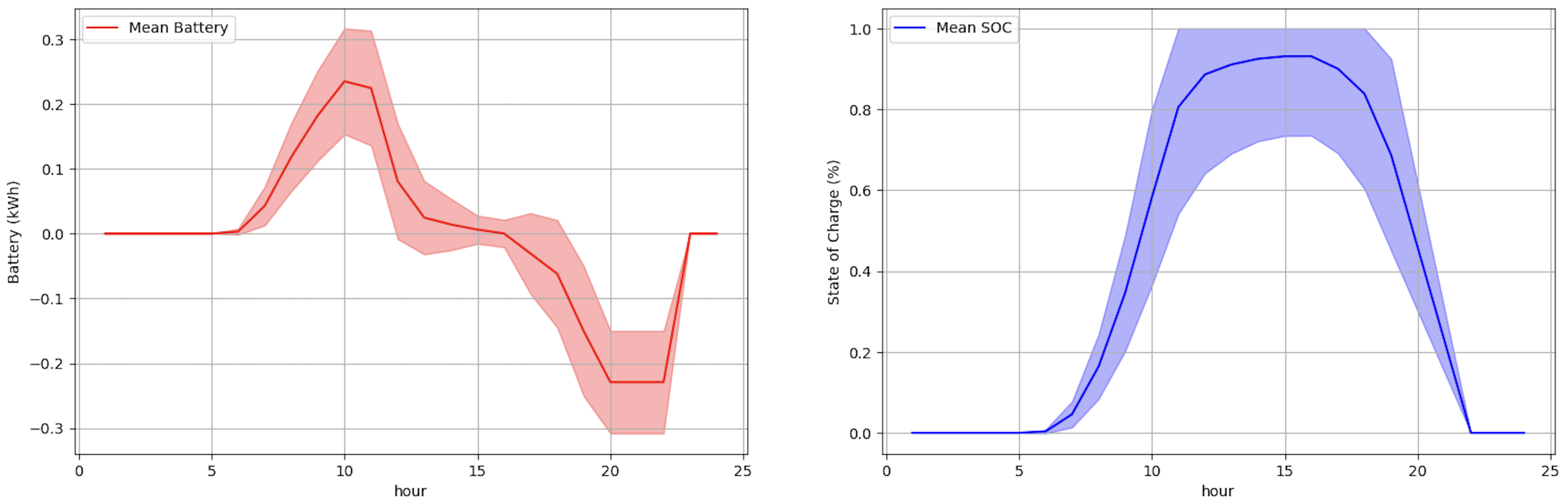}
    \caption{Optimal Actions and State of Charge: Mean and variance throughout the day, considering a complete year, for a building. By design, we know the optimal policy that must yield zero emissions and cost beforehand.}
    \label{fig:eg_opt_act_soc}
\end{figure}

\section{Synthetic Data}

For this research, we decided to generate our data as it allowed us to explain better the policies we learned. Although using accurate data is always desirable for ML, there is low availability, especially regarding energy usage patterns, which concerns privacy. For example, the CityLearn environment constructs its 2022 scenario using data from 17 Sierra Crest Zero Net Energy community buildings from another study about ToU strategies \cite{narayanamurthy2016grid}. They couldn't access all the accurate data details, so they had to generate data that emulated the behavior based on some prior information about the ToU strategies the households might have used. Research about microgrid control systems is not particularly affected by lack of data, as prior information about consumption behaviors, typical weather conditions given a location, seasonal changes, and factors influencing energy use are accessible. Synthetic data could represent typical scenarios an agent may face and have the advantage of creating relevant evaluation scenarios by adding noise or making a convex combination of seen data.

\begin{figure} [hbt]
    \centering
    \subfigure[Simplified Dataset]{\label{fig:simple_solar_load}\includegraphics[width=0.8\textwidth]{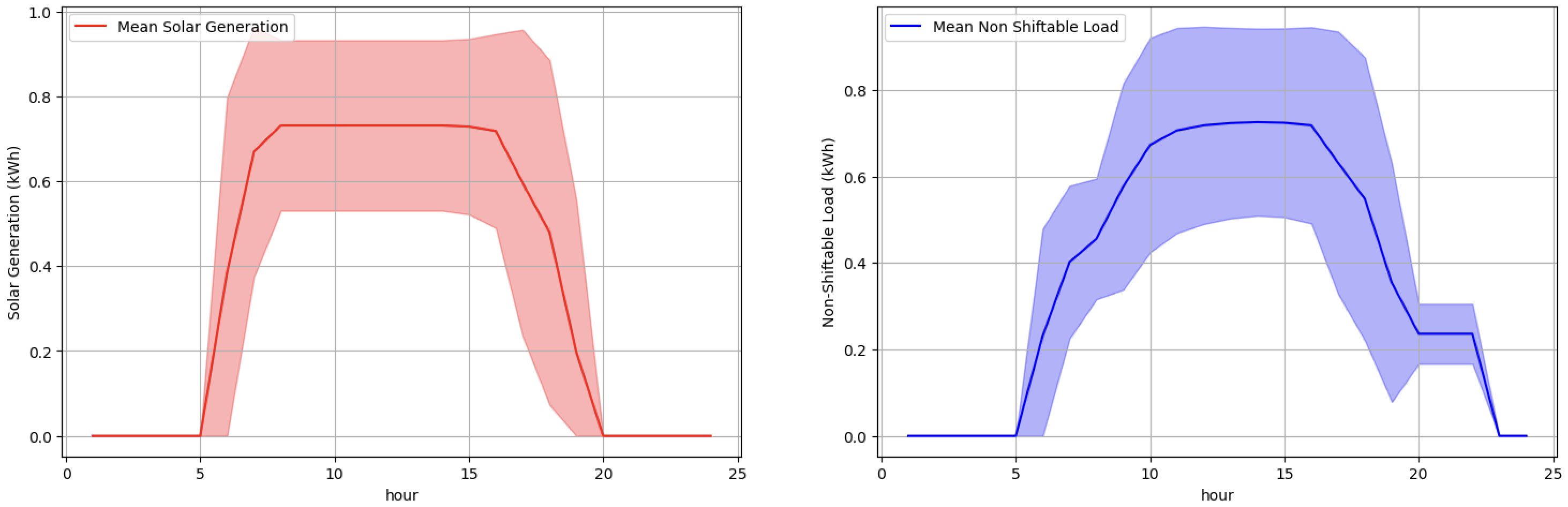}}
    \subfigure[Normal Dataset]{\label{fig:normal_solar_load}\includegraphics[width=0.8\textwidth]{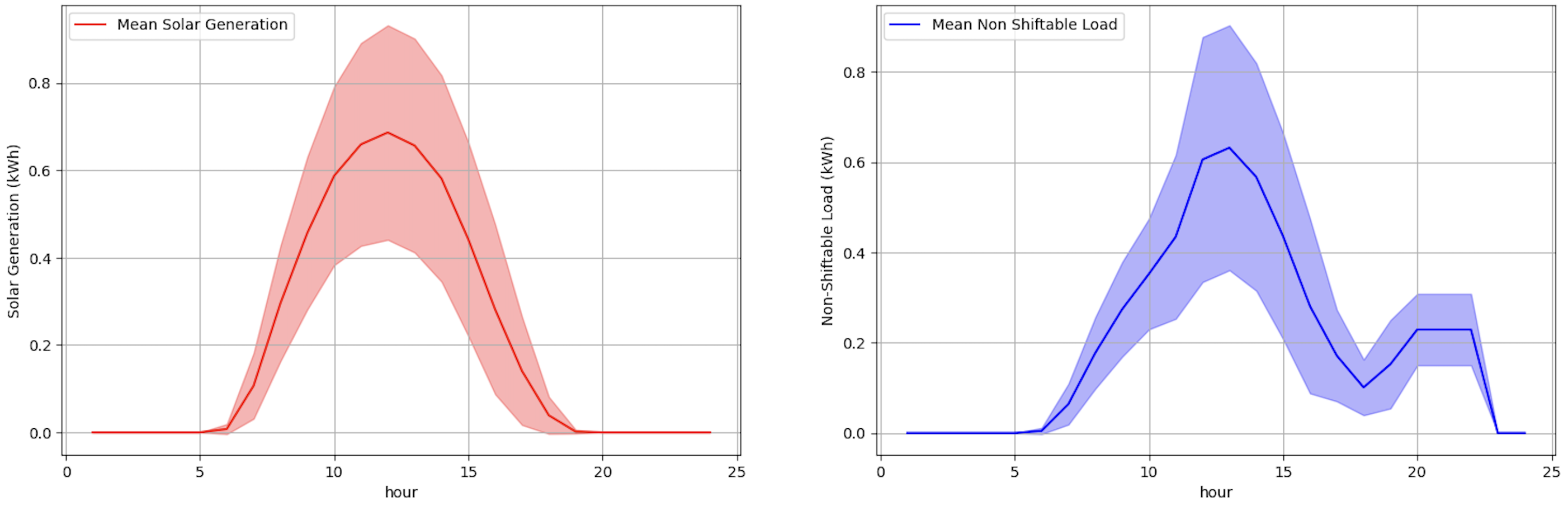}}
    \caption{Sample of the Mean and Variance of Solar Generation and Non-Shiftable Load over 24 hours for a year of data for one of the buildings.}
    \label{fig:dataset_samples_solar_load}
\end{figure}

\begin{figure} [hbt]
\centering
    \subfigure[Original scenario]{\label{fig:original_pricing}\includegraphics[width=0.46\textwidth]{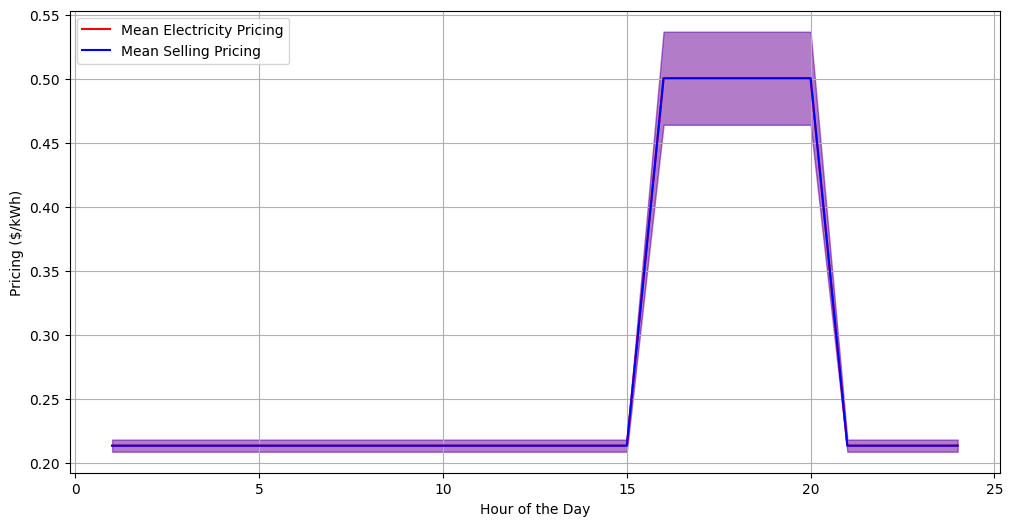}}
    \hfill
    \subfigure[Our scenario]{\label{fig:our_pricing}\includegraphics[width=0.46\textwidth]{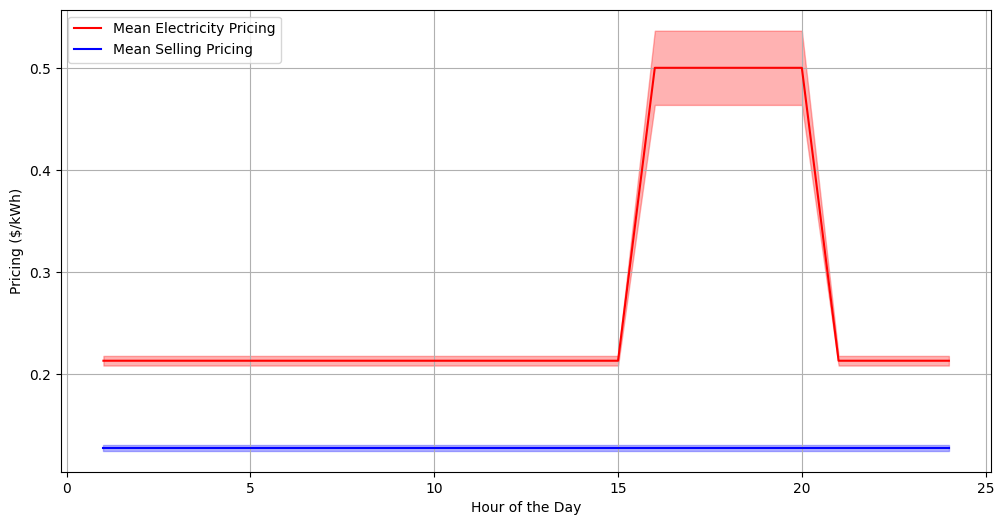}}
    \caption{Mean and Variance of Electricity Pricing (ToU) and the Selling Pricing over 24 hours for a year of data.}
    \label{fig:dataset_pricing}
\end{figure}

We generated our synthetic data by altering the Non-Shiftable Load time series using a simple rule that defines the policy we want to learn: the households have enough energy and battery capacity to serve their demand when there is solar energy and to store what is necessary when there is no sun. By definition, our scenario leads to net-zero energy consumption. Figure \ref{fig:dataset_samples_solar_load} shows how two versions of the data we generated, in the case of \ref{fig:simple_solar_load}, we simplified the solar profile by making it constant through the sunny hours. For \ref{fig:normal_solar_load}, we kept the solar profile as in the original reference case from CityLearn. By simplifying these profiles, we intended to reduce the state space for the policy while still keeping it challenging to learn. Another critical difference in our scenario is the pricing for energy. Our reference scenario from the CityLearn environment used the real pricing time series used in the Zero Crest Community Study. Still, for the environment implementation, they defined that the selling price is the same tariff whenever a household sends back energy to the grid. By inspecting \ref{fig:original_pricing}, it is easy to notice that a policy optimizing exclusively by the energy cost would charge as much as possible in the low-price hours to sell energy when it is almost twice the buying price. In our scenario, we defined a margin of 40\% of the lowest daily electricity pricing to ensure we don't learn policies that bypass the weighting factor $W_G$ by finding profitable energy trades.

\begin{table}[h]
    \centering
    \caption{ RL case for \textbf{agent} $\mathbf{i}$: detail of the state composition with the units of each feature, action definition, and reward design.}
    \begin{tabular}{@{}lrlc@{}}

        \toprule
        \multicolumn{4}{c}{\textbf{State: $\mathbf{s_{t,i}}$}} \\
        \midrule

        Hour & $t$ & $\in~\{1, 2, \ldots, 23\}$ & $-$ \\
        \hdashline[.4pt/1pt]
        
        Day Type & $d$ & $\in~\{1, 2, \ldots, 7\}$ & $-$ \\
        \hdashline[.4pt/1pt]
        
        Solar Generation & $E_{t,i}^{\text{solar}}$ & $\in~\mathbb{R}_{\geq 0}$ & $kWh$  \\
        \hdashline[.4pt/1pt]
        
        Non-Shiftable Load & $E_{t,i}^{\text{load}}$ & $\in~\mathbb{R}_{\geq 0}$ & $kWh$ \\
        \hdashline[.4pt/1pt]
        
        Electrical Storage SOC & $B_{t,i}^{\text{soc}}$ & $\in~[0, 1]$ & $\%$ \\
        \hdashline[.4pt/1pt]
        
        Net Electricity Consumption & $E_{t,i}^{\text{net}}$ & $\in~\mathbb{R}$ & $kWh$ \\
        \hdashline[.4pt/1pt]
        
        Predicted Load (4h) & $E_{t,i}^{\text{load+4}}$ & $\in~\mathbb{R}_{\geq 0}$ & $kWh$ \\
        \hdashline[.4pt/1pt]
        
        Predicted Load (6h) & $E_{t,i}^{\text{load+6}}$ & $\in~\mathbb{R}_{\geq 0}$ & $kWh$ \\
        \hdashline[.4pt/1pt]
        
        Predicted Load (12h) & $E_{t,i}^{\text{load+12}}$ & $\in~\mathbb{R}_{\geq 0}$ & $kWh$ \\
        \hdashline[.4pt/1pt]
        
        Predicted Load (24h) & $E_{t,i}^{\text{load+24}}$ & $\in~\mathbb{R}_{\geq 0}$ & $kWh$ \\
        \hdashline[.4pt/1pt]
        
        Direct Solar Irradiance & $I_{t,i}^{\text{direct}}$ & $\in~\mathbb{R}$ & $W/m^2$ \\
        \hdashline[.4pt/1pt]
        
        Predicted Solar Irradiance (4h) & $I_{t,i}^{\text{direct+4}}$ & $\in~\mathbb{R}_{\geq 0}$ & $W/m^2$ \\
        \hdashline[.4pt/1pt]
        
        Predicted Solar Irradiance (6h) & $I_{t,i}^{\text{direct+6}}$ & $\in~\mathbb{R}_{\geq 0}$ & $W/m^2$ \\
        \hdashline[.4pt/1pt]
        
        Predicted Solar Irradiance (12h) & $I_{t,i}^{\text{direct+12}}$ & $\in~\mathbb{R}_{\geq 0}$ & $W/m^2$ \\
        \hdashline[.4pt/1pt]
        
        Predicted Solar Irradiance (24h) & $I_{t,i}^{\text{direct+24}}$ & $\in~\mathbb{R}_{\geq 0}$ & $W/m^2$ \\
        \hdashline[.4pt/1pt]
        
        Electricity Pricing & $R_{t,i}^{\text{buy}}$ & $\in~\mathbb{R}_{\geq 0}$ & $\text{USD}/kWh$ \\
        \hdashline[.4pt/1pt]
        
        Selling Pricing & $R_{t,i}^{\text{sell}}$ & $\in~\mathbb{R}_{\geq 0}$ & $\text{USD}/kWh$ \\
        \hdashline[.4pt/1pt]

        Carbon Intensity & $R_{t,i}^{\text{carbon}}$ & $\in~\mathbb{R}_{\geq 0}$ & $\text{kgCO}_2/kWh$ \\
        
        \midrule
        \multicolumn{4}{c}{\textbf{Other Relevant Values}} \\
        \midrule

        Battery Capacity & $B_{i}^{\text{cap}}$ & $\in~\mathbb{R}_{\geq 0}$ & $kWh$\\
        \hdashline[.4pt/1pt]
        
        Energy Cost Weight & $W_{C}$ & $\in~[0, 1]$ & $-$\\
        \hdashline[.4pt/1pt]
        
        Energy Emissions Weight & $W_{G}$ & $\in~[0, 1]$ & $-$\\
        \hdashline[.4pt/1pt]

        \multicolumn{4}{c}{\footnotesize \textit{Note:} $W_C + W_G = 1$} \\
        
        \midrule
        \multicolumn{4}{c}{\textbf{Action: $\mathbf{a_{t,i}}$}} \\
        \midrule

        Battery charge/discharge setpoint & $B^{\text{action}}_{t,i}$ & $\in~[-1, 1]$ & $\%$\\

        \midrule
        \multicolumn{4}{c}{\textbf{Reward: $\mathbf{r_{t,i}}$}} \\
        \midrule

        \multicolumn{4}{c}{$P_{i,t} = B_{i}^{\text{cap}} * \begin{cases}
            \max\left\{0, B^{\text{action}}_{t-1,i} - (1 - B_{t-1,i}^{\text{soc}})\right\}
            &~~\text{if}~~
            B^{\text{action}}_{t-1,i} \geq 0
            \\
            \max\left\{0, |B^{\text{action}}_{t-1,i}| - B_{t-1,i}^{\text{soc}}\right\}
            &~~\text{o.w.}
        \end{cases}$}\vspace{1mm} \\ 
        \hdashline[.4pt/1pt]

        \multicolumn{4}{c}{$C_{i,t} = 
        \max\left\{
            0, E_{t,i}^{\text{net}} + P_{i,t}
        \right\} * R_{t,i}^{\text{buy}}
        + 
        \min\left\{
            0, E_{t,i}^{\text{net}} + P_{i,t}
        \right\} * R_{t,i}^{\text{sell}}
        $}\vspace{1mm} \\ 
        \hdashline[.4pt/1pt]

        \multicolumn{4}{c}{$G_{i,t} = 
        \max\left\{
            0, E_{t,i}^{\text{net}} + P_{i,t}
        \right\} * R_{t,i}^{\text{carbon}}
        $}\vspace{1mm} \\ 
        \hdashline[.4pt/1pt]
        
        \multicolumn{4}{c}{$
        r_{t,i} = - \left( 
            C_{i,t} * W_C + G_{i,t} * W_G
        \right)
        $} \\

        \bottomrule
    \end{tabular}
    \label{tab:rl_case}
\end{table}

\section{Hyperparameters}

We perform a hyperparameter search with Wandb sweeps \cite{wandb}, using Bayesian search to find the best configurations for PPO and TRPO with the following final configurations:

\subsubsection{Quantitative Results}

\begin{figure} [hbt]
    \centering
    \subfigure[2B Eval. Energy Cost]{\label{fig:best_2b_ppo_eval_cost}\includegraphics[width=0.45\textwidth]{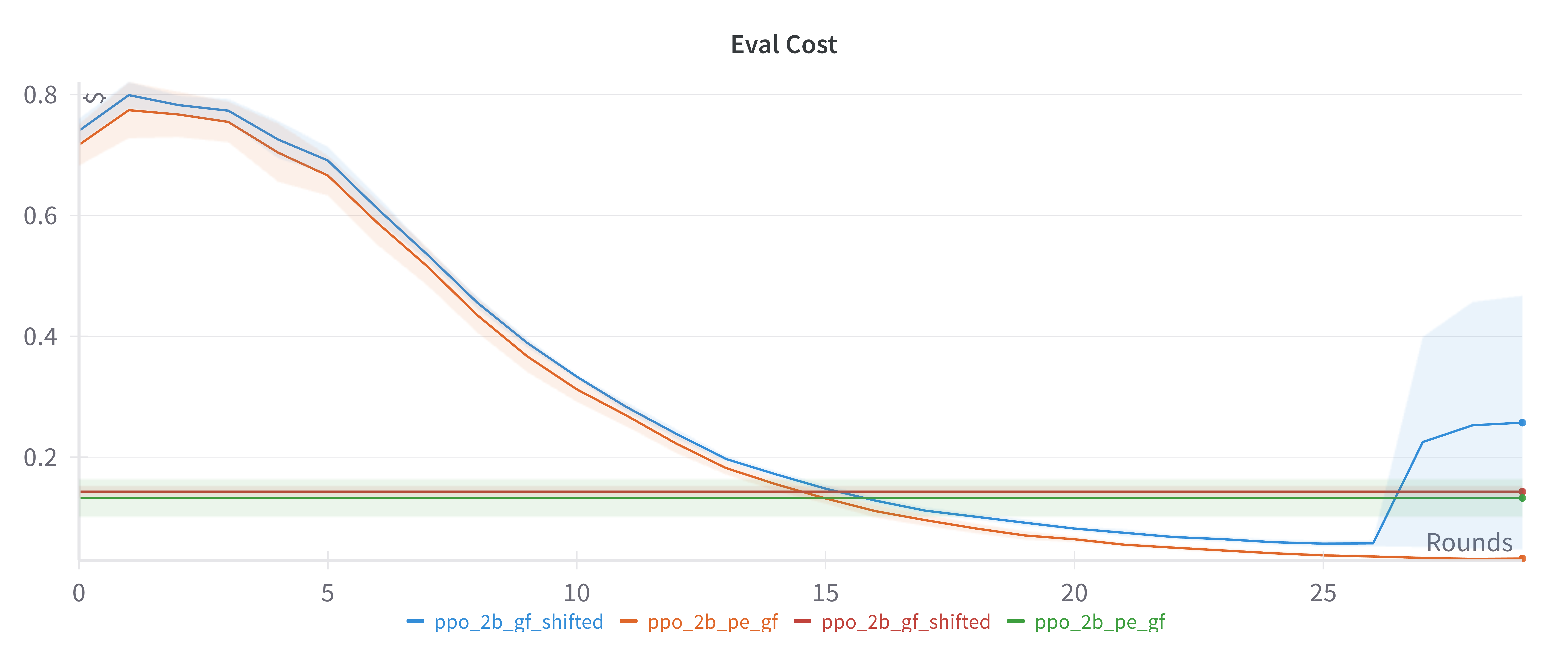}}
    \subfigure[2B Eval. Emissions]{\label{fig:best_2b_ppo_eval_emissions}\includegraphics[width=0.45\textwidth]{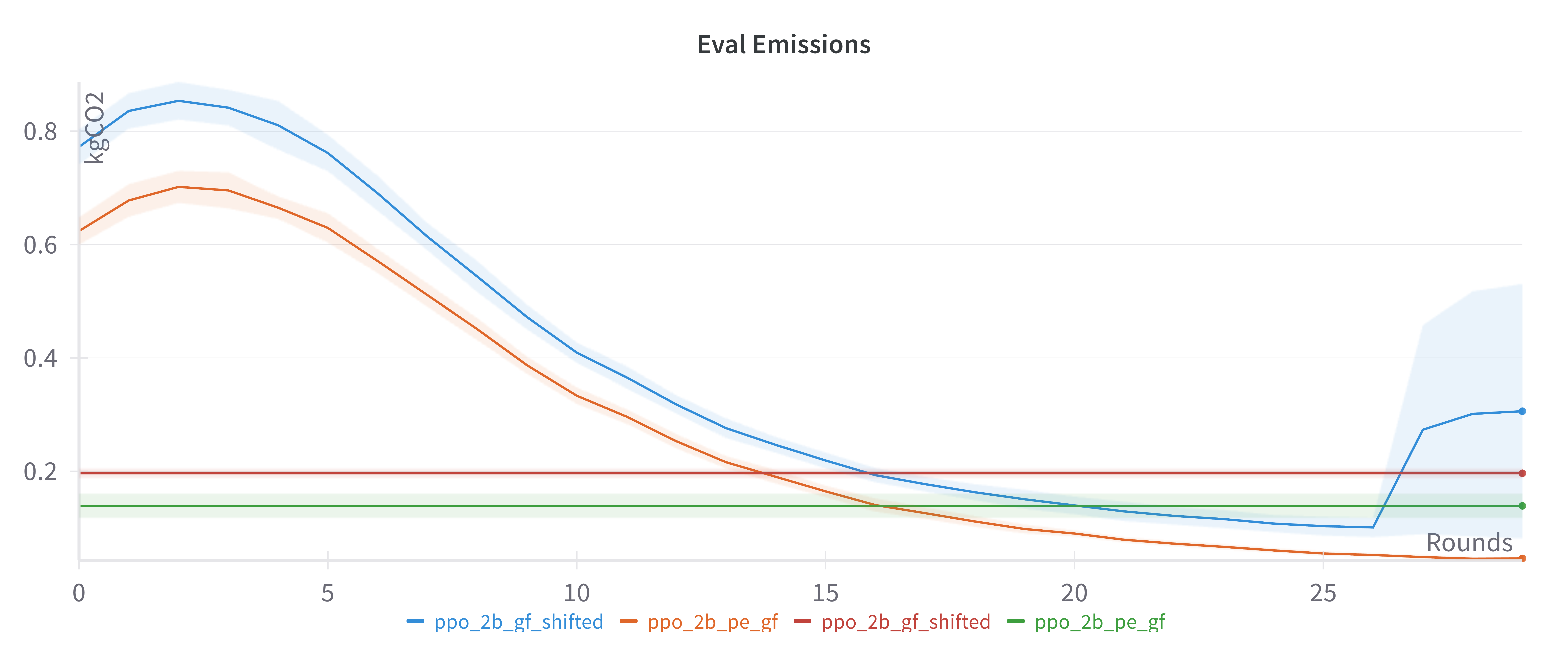}}
    \subfigure[5B Eval. Energy Cost]{\label{fig:best_5b_ppo_eval_cost}\includegraphics[width=0.45\textwidth]{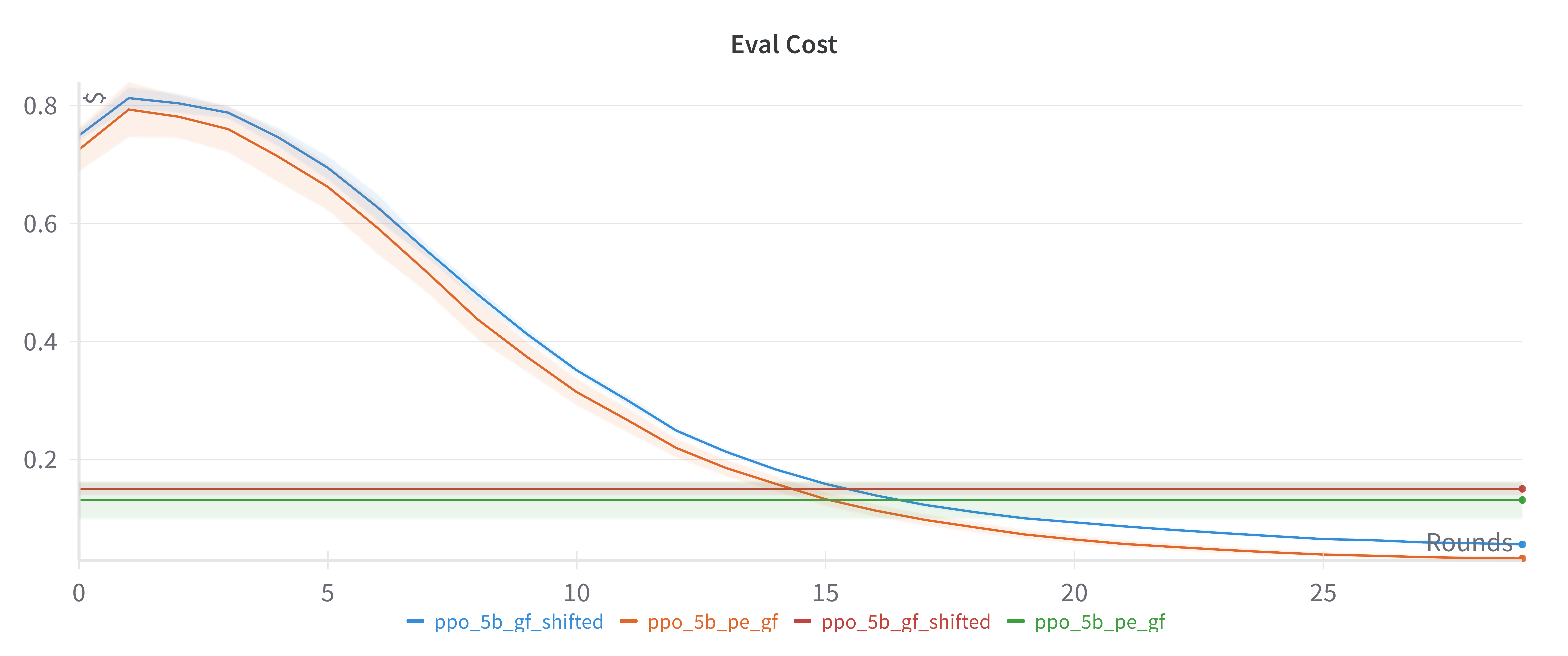}}
    \subfigure[5B Eval. Emissions]{\label{fig:best_5b_ppo_eval_emissions}\includegraphics[width=0.45\textwidth]{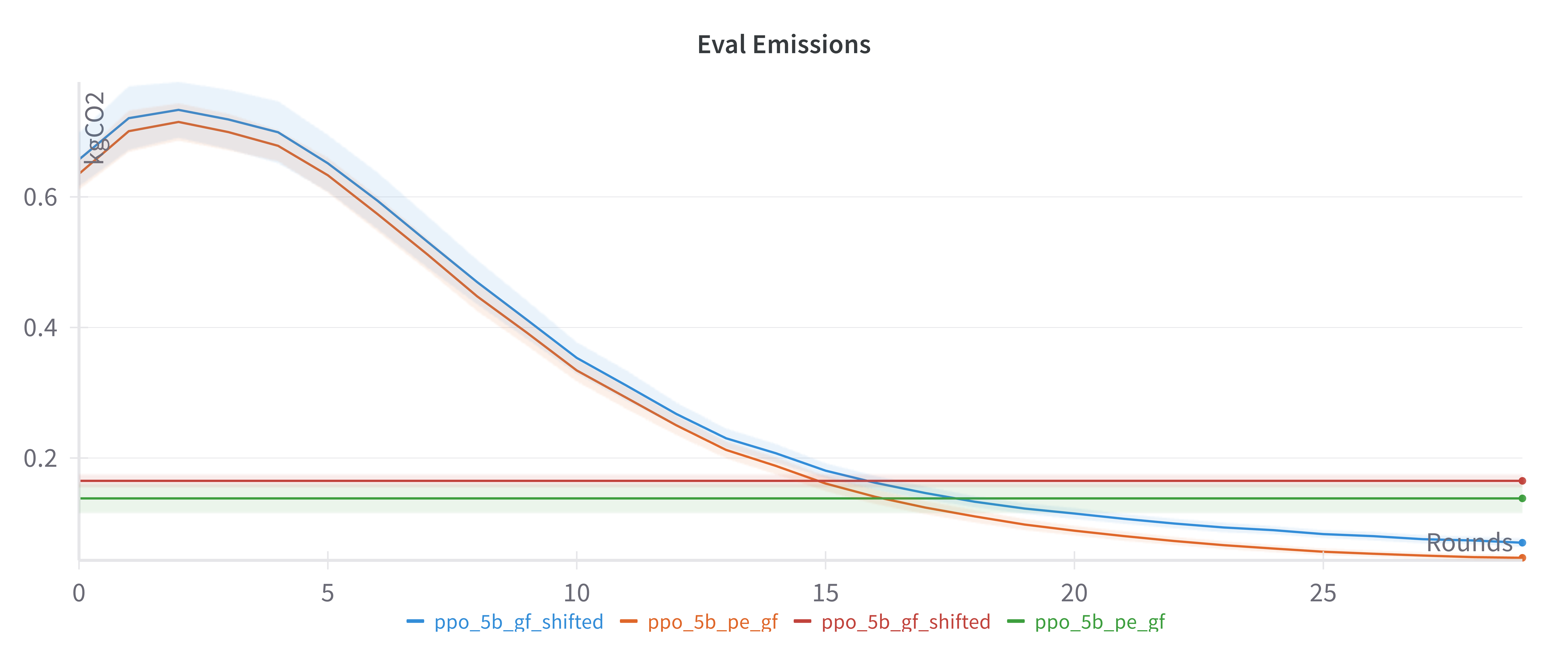}}
    \caption{\textbf{Best PPO policies:} Plots show mean with an envelope of one standard deviation computed from the five different seeds.}
    \label{fig:best_ppo}
\end{figure}

We inspected the best-performing policies, considering the eval. reward, according to Tables \ref{tab:exp_results_2b} and \ref{tab:exp_results_5b} to understand if we accomplished the objective of beating the baseline policy, which is not using a battery. In the case of PPO, we can notice in \ref{fig:best_ppo} that it was the case. However, we can see an effect we mentioned before in Figures  \ref{fig:best_2b_ppo_eval_cost} and \ref{fig:best_2b_ppo_eval_emissions}: extending the training to reduce the optimality gap destabilizes the training, which is something that happens to gradient-based optimizers with a fixed learning rate.

Increasing the number of buildings makes it harder for the agents to approach net-zero energy and training with the shifted version of the dataset, something expected. A first inspection of the results for the five-building environment might give the impression that the training process was more stable, but the reason must be that the policy didn't get as close to net zero as in the case of two buildings; the instability in the training happens in the vicinity of the optimal solution.

\begin{figure} [hbt]
    \centering
    \subfigure[2B Eval. Energy Cost]{\label{fig:best_2b_trpo_eval_cost}\includegraphics[width=0.45\textwidth]{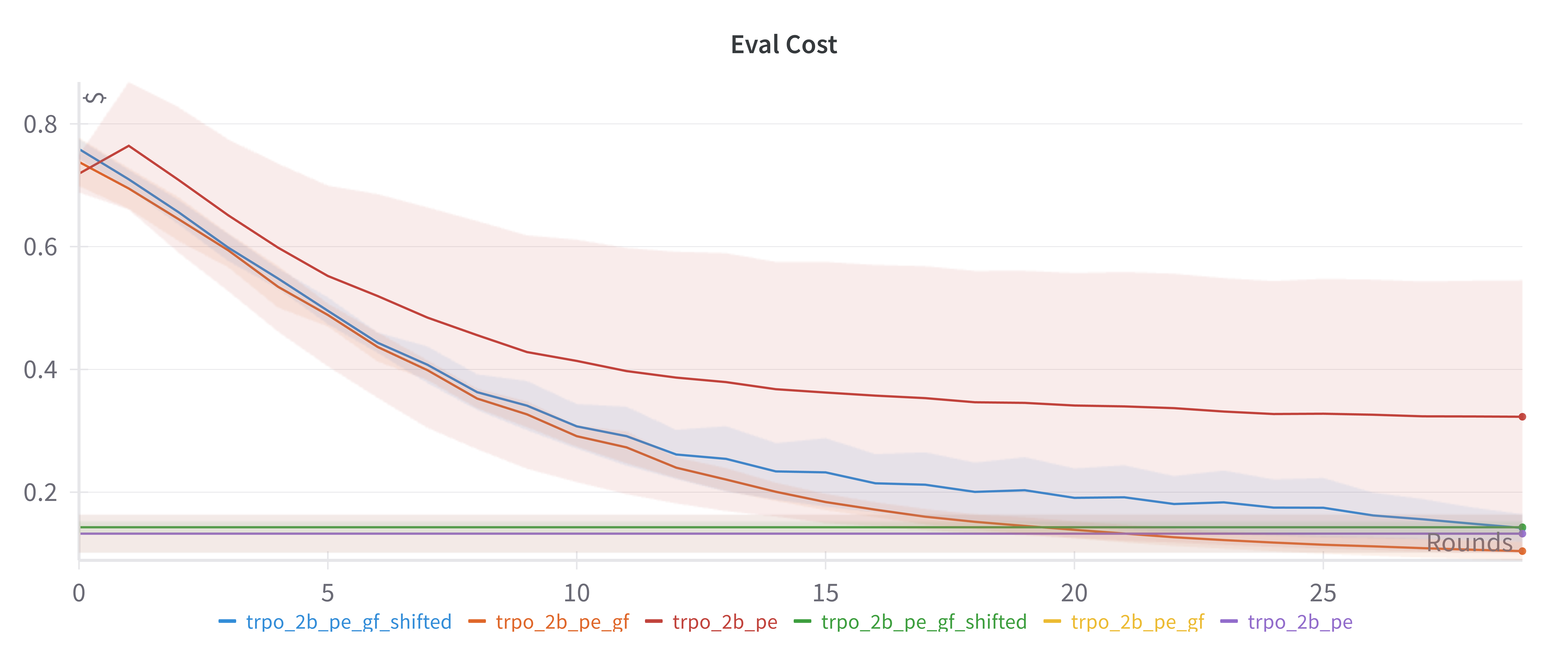}}
    \subfigure[2B Eval. Emissions]{\label{fig:best_2b_trpo_eval_emissions}\includegraphics[width=0.45\textwidth]{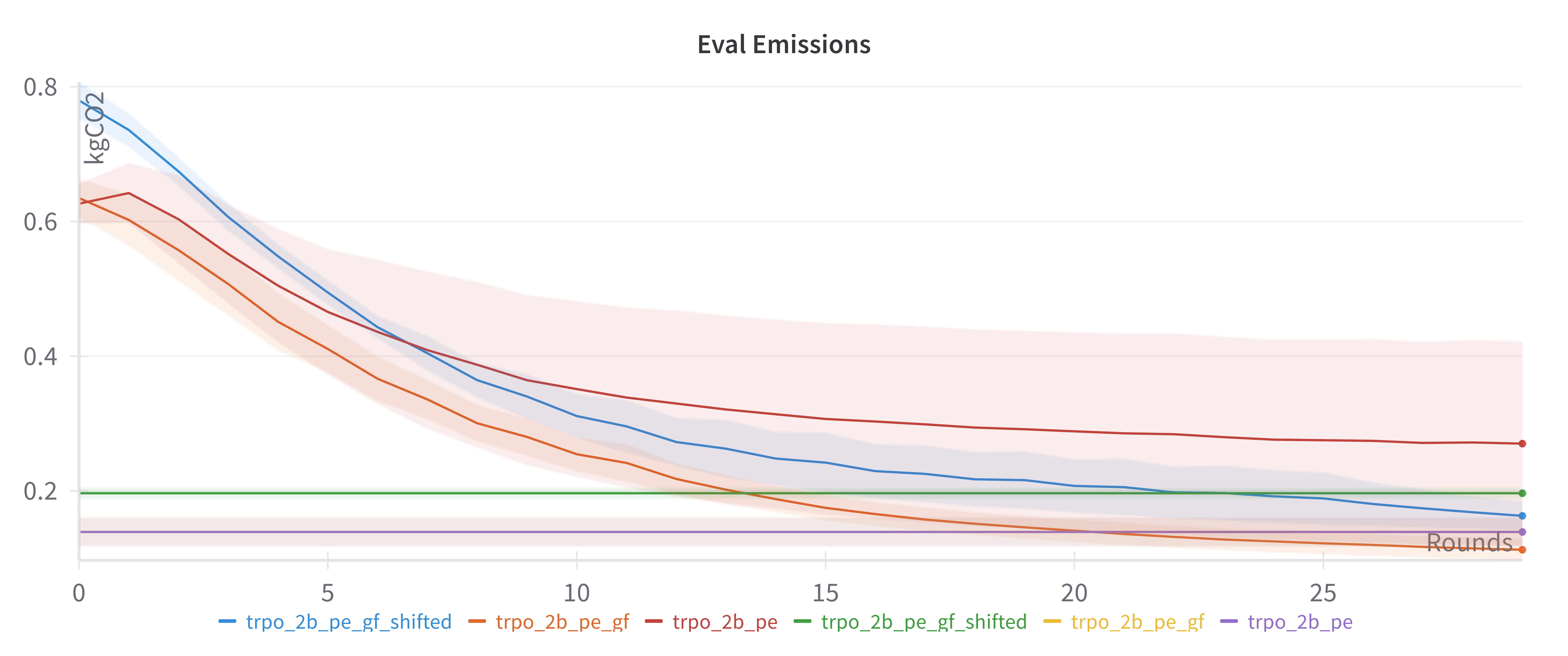}}
    \subfigure[5B Eval. Energy Cost]{\label{fig:best_5b_trpo_eval_cost}\includegraphics[width=0.45\textwidth]{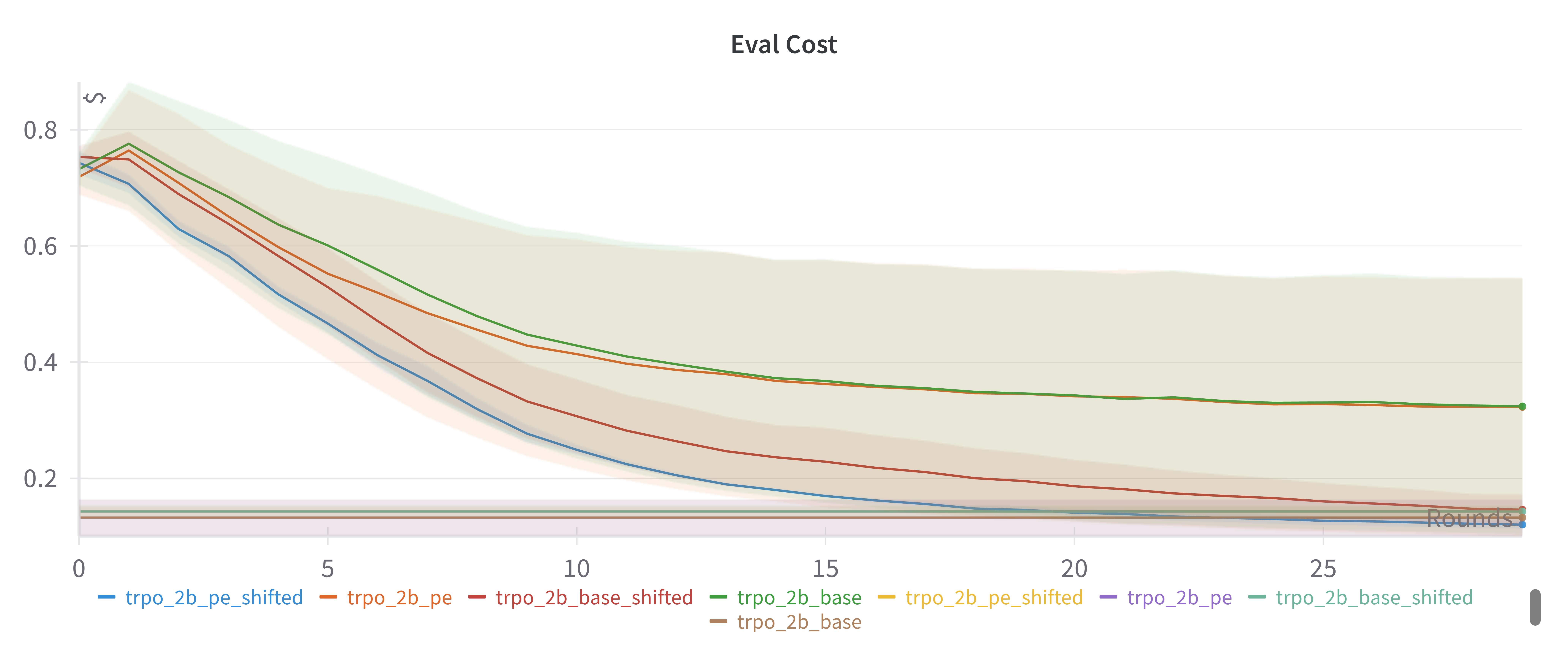}}
    \subfigure[5B Eval. Emissions]{\label{fig:best_5b_trpo_eval_emissions}\includegraphics[width=0.45\textwidth]{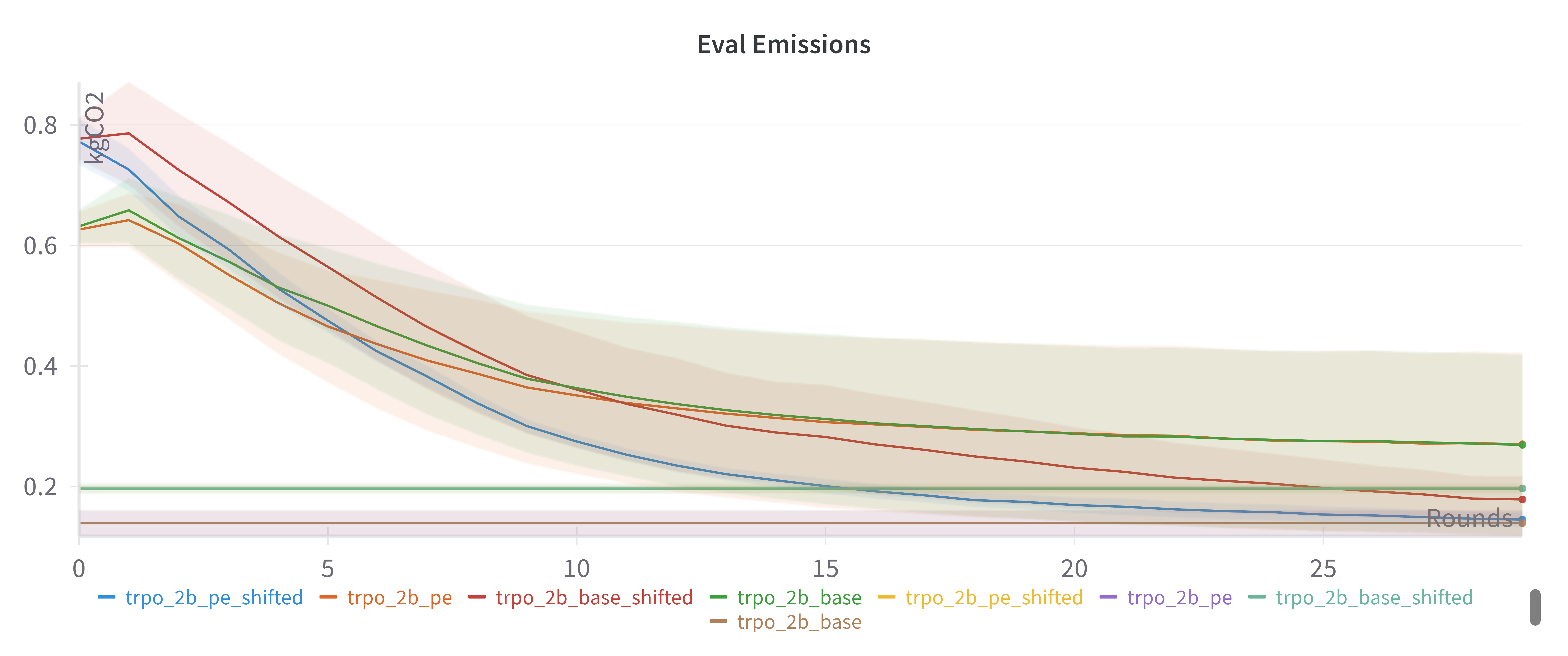}}    
    \caption{\textbf{Best TRPO policies:} Plots show mean with an envelope of one standard deviation computed from the five different seeds.}
    \label{fig:best_trpo}
\end{figure}

For the best-performing policies using TRPO, we can notice how the low returns reflect high variance among runs. However, some configurations (model with personal encoding and grouping) limited the variance. They beat the baseline on each run, which is a highlight considering the lack of tuning. That event was present in both environments.

\begin{figure} [hbt]
    \centering
    
    \subfigure[PPO: B0 Policy]{\label{fig:ppo_2b_best_actions_0}\includegraphics[width=0.23\textwidth]{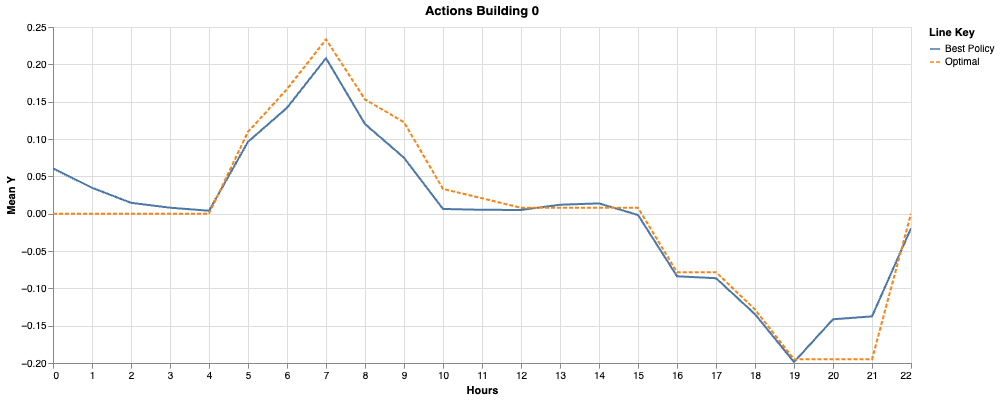}}
    \subfigure[PPO: B0 SoC]{\label{fig:ppo_2b_best_soc_0}\includegraphics[width=0.23\textwidth]{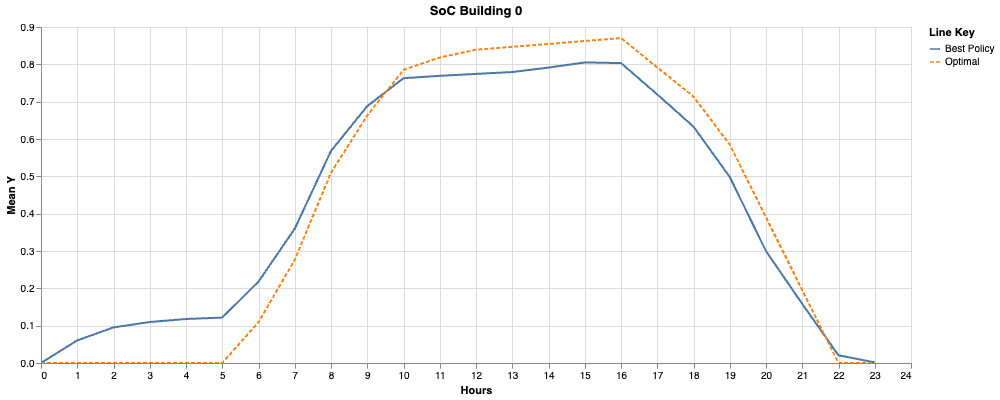}}
    \subfigure[PPO: B1 Policy]{\label{fig:ppo_2b_best_actions_1}\includegraphics[width=0.23\textwidth]{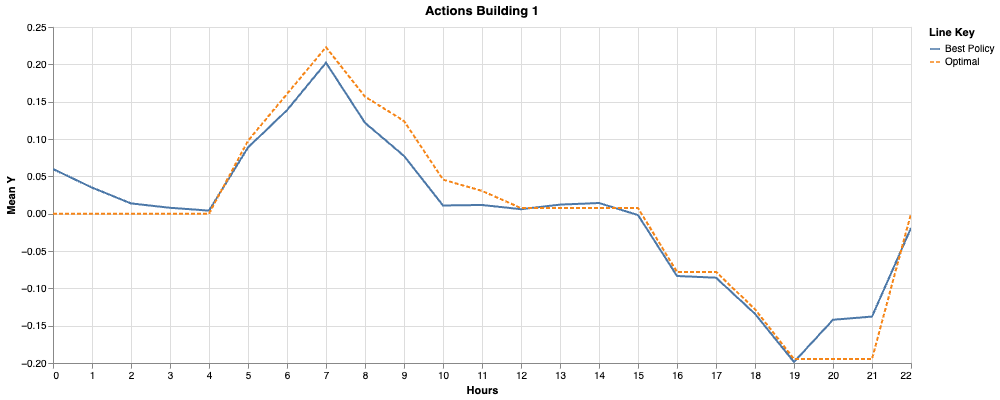}}
    \subfigure[PPO: B1 SoC]{\label{fig:ppo_2b_best_soc_1}\includegraphics[width=0.23\textwidth]{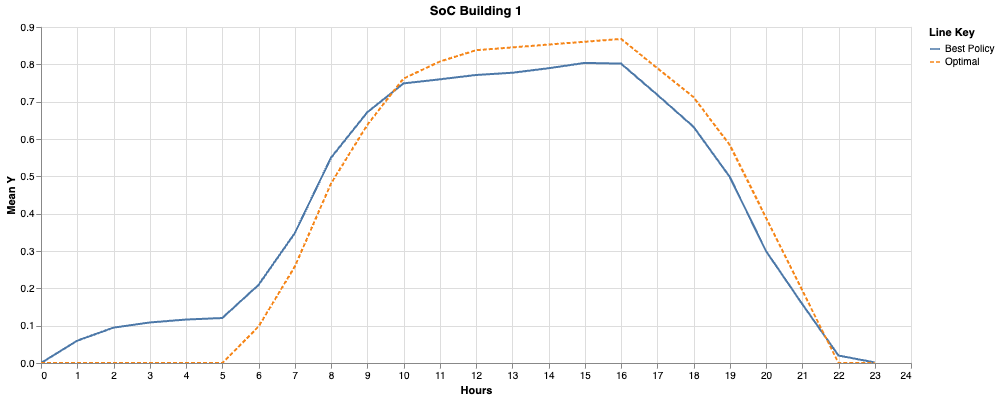}}
    \\
    \subfigure[TRPO: B0 Policy]{\label{fig:trpo_2b_best_actions_0}\includegraphics[width=0.23\textwidth]{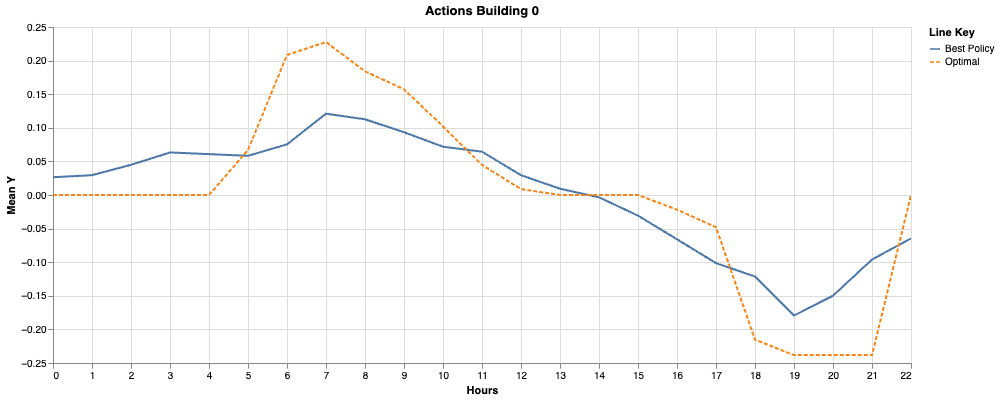}}
    \subfigure[TRPO: B0 SoC]{\label{fig:trpo_2b_best_soc_0}\includegraphics[width=0.23\textwidth]{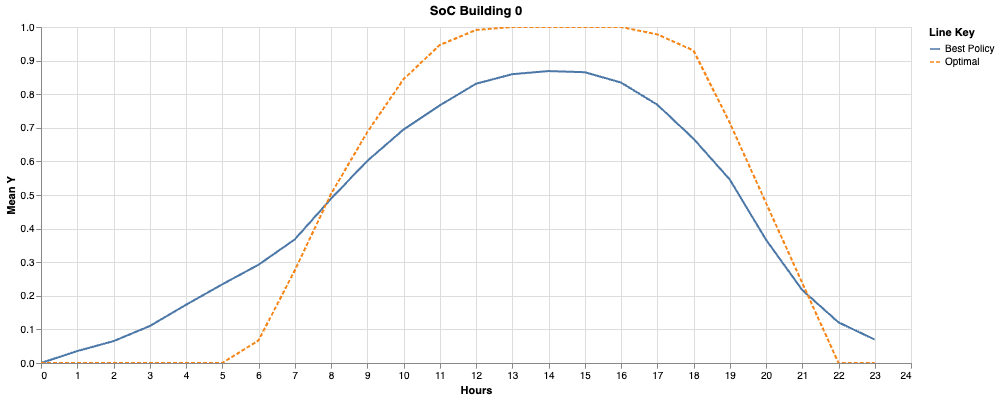}}
    \subfigure[TRPO: B1 Policy]{\label{fig:trpo_2b_best_actions_1}\includegraphics[width=0.23\textwidth]{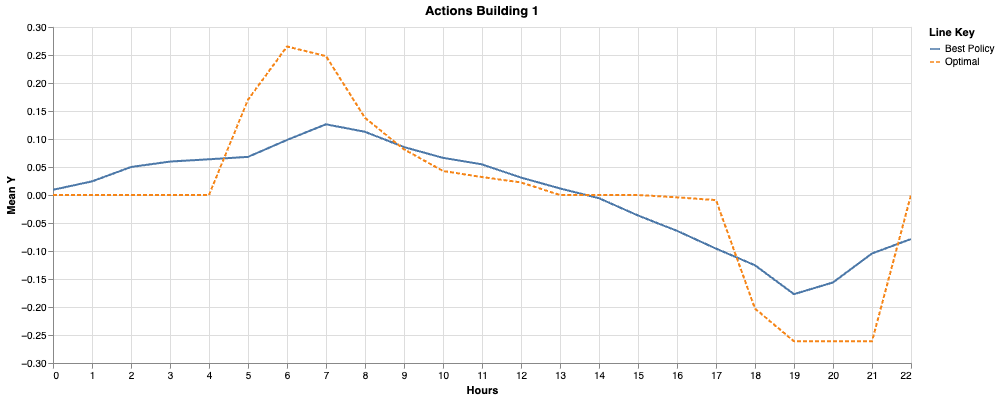}}
    \subfigure[TRPO: B1 SoC]{\label{fig:trpo_2b_best_soc_1}\includegraphics[width=0.23\textwidth]{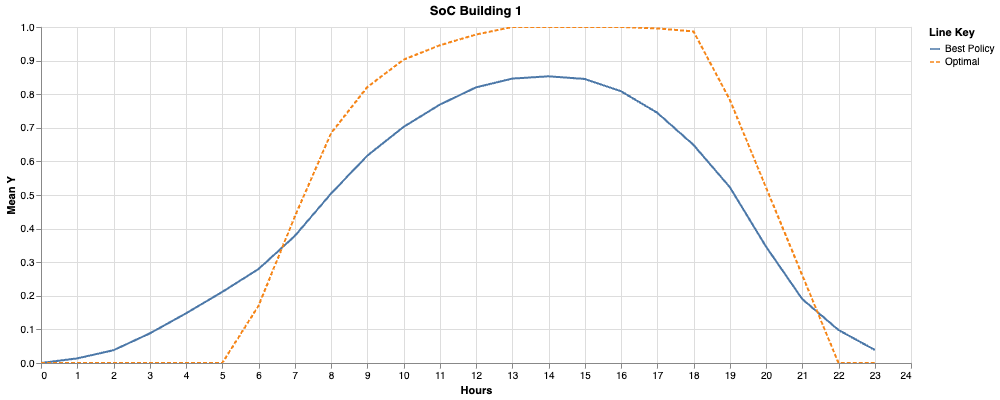}}
    
    \caption{\textbf{Best policies for the 2-building environment:} Comparison of learned policy against the optimal policy, averaged for the five seeds.}
    \label{fig:best_policies}
\end{figure}

In Figure \ref{fig:best_policies}, we notice that PPO gets very close to the optimal policy, but a few misalignments explain why it couldn't reach net zero. From the best TRPO policy, we can highlight that although it's far from optimal, it is evident that it learned a pattern, and that although sub-optimal, it's still a policy that beats the baseline.

\section{Results Summary}

When analyzing the results, it is essential to consider that the PPO approach was tuned and used the best set of hyperparameters we found, while TRPO was not tuned. Both techniques rely on the same theoretical base: both fall under the policy gradient-based methods of RL, both aim to optimize a surrogate function for the return based on the Performance Difference Lemma (PDL) proposed in \cite{kakade2002approximately}, both aim to do conservative updates to the policy parameters to ensure a stable learning process using the Kullback-Leibler (KL) Divergence between the iterations of the policy's distribution. The main difference lies in removing the use of second-order information in the optimization process (TRPO), replacing it with a clipped version (PPO). Each approach has its advantages and disadvantages. We faced them during the training process, and it also reflected in the results. For instance, PPO faced challenges when trying to reduce the optimality gap in the vicinity of the solution, and adding more iterations destabilized the training (even using a small learning rate). TRPO suffered when facing a nonconvenient initialization, which is a common problem in second-order methods, limiting its capability of exploiting sampling (more trajectories would lead to a better Hessian estimation) to optimize in fewer iterations with an adaptive learning rate.



\end{document}